\documentclass[10pt]{article}
\usepackage{fancyhdr}
\usepackage{extramarks}
\usepackage{amsmath}
\usepackage{amsthm}
\usepackage[utf8]{inputenc}   % arXiv/pdflatex-friendly
\usepackage[T1]{fontenc}
\usepackage{newunicodechar}   % map a few Unicode run-ins to TeX
 \usepackage{dblfloatfix} % improves placement of figure* in two-column docs
\usepackage{float}

\usepackage{booktabs}
\usepackage{tabularx}

\renewcommand{\arraystretch}{1.2}  % slightly wider row spacing

\newunicodechar{α}{$\alpha$}
\newunicodechar{β}{$\beta$}
\newunicodechar{γ}{$\gamma$}
\newunicodechar{κ}{$\kappa$}
\newunicodechar{δ}{$\delta$}

\newunicodechar{–}{--}       % en dash
\newunicodechar{—}{---}      % em dash
\newunicodechar{’}{'}        % apostrophe
\newunicodechar{±}{$\pm$}

\newunicodechar{­}{}
\usepackage{amsfonts}
\usepackage{siunitx}
\usepackage{tikz}
\usepackage[plain]{algorithm}
\usepackage{algpseudocode}
\usepackage{multirow}
\usepackage{booktabs}
\usepackage{graphicx}
\usepackage{subfigure}
\usepackage[margin=1in]{geometry}
\usepackage{booktabs}
\usepackage{makecell}
\usepackage{array} 
\usepackage[colorlinks,linkcolor=black,anchorcolor=black,citecolor=black,urlcolor=blue]{hyperref}
\usepackage{hyphenat}
\usepackage{amsmath,bm}
\usepackage{booktabs}
\usepackage{mathtools}
\usepackage{amssymb}
\usepackage{tikz-cd}
\usepackage{caption}
\usepackage{capt-of}
\usepackage{mciteplus}
\usepackage{cite}
\usepackage{mathrsfs}
\usepackage[title,titletoc,toc]{appendix}
\usepackage{xr}
\usepackage{parskip}
\usepackage{soul}
\usepackage{textcomp}
\usepackage[colaction]{multicol}
\usepackage[switch]{lineno}
\usepackage{lipsum}
\usepackage{etoolbox}
\usepackage{longtable}
\usepackage{array}
\usepackage{tablefootnote}
\usepackage{ragged2e}
\usepackage{soul}
\newcolumntype{C}[1]{>{\centering\arraybackslash}p{#1}}
\usetikzlibrary{automata,positioning}
\usepackage{float}

\usetikzlibrary{automata,positioning}

\newtheorem{remark}{Remark}[section]

\def\rank{\mbox{\rm rank\,}}

\def\Gr{{\mbox{\bf Gr}}}

\usepackage{amsthm}  %  
\theoremstyle{definition}
\newtheorem{definition}{Definition}[section]

\usepackage{xr}
\begin{document}
	\title{Multiscale Grassmann Manifolds for Single-Cell Data Analysis}
    %\title{Precision Drug Repurposing for Alzheimer's Disease Enabled via Persistent Sheaf Laplacian Analysis of Gene Co-Expression Networks }
    
	\author{Xiang Xiang Wang$^{1}$, Sean Cottrell$^{1,2}$, and Guo-Wei Wei$^{1,3,4}$\footnote{
			Corresponding author.		Email: weig@msu.edu} \\% Author name
		\\
		$^1$ Department of Mathematics, \\
		Michigan State University, East Lansing, MI 48824, USA.\\
        $^2$ Department of Computational Mathematics, Science, and Engineering, \\
        Michigan State University, East Lansing, MI 48824, USA. \\	
        $^3$ Department of Biochemistry and Molecular Biology,\\
		Michigan State University, East Lansing, MI 48824, USA.  \\
		$^4$ Department of Electrical and Computer Engineering,\\
		Michigan State University, East Lansing, MI 48824, USA. \\
        \\
	}
	\date{\today} % Date for the report
	
	\maketitle
	
\begin{abstract}
Single-cell data analysis seeks to characterize cellular heterogeneity based on high-dimensional gene expression profiles. 
Conventional approaches represent each cell as a vector in Euclidean space, which limits their ability to capture intrinsic correlations and multiscale geometric structures. 
We propose a multiscale framework based on Grassmann manifolds that integrates machine learning with subspace geometry for single-cell data analysis. 
By generating embeddings under multiple representation scales, the framework combines their features from different geometric views into a unified    Grassmann manifold. 
A power-based scale sampling function is introduced to control the selection of scales and balance information across resolutions. 
Experiments on nine benchmark single-cell RNA-seq datasets demonstrate that the proposed approach effectively preserves  meaningful structures and provides stable clustering performance, particularly for small to medium-sized datasets. 
These results suggest that Grassmann manifolds offer a coherent and informative foundation for analyzing single cell data.
\end{abstract}

\noindent\textbf{Keywords:} Grassmann manifolds, multiscale, subspace representation,  single-cell data analysis

	%{\setcounter{tocdepth}{4} \tableofcontents}
	%\setcounter{page}{1}

\section{Introduction}

Single-cell sequencing has become an important tool for studying cellular heterogeneity and elucidating biological mechanisms at the single-cell level~\cite{choi2019dissecting}. 
Modern single-cell sequencing technologies measure the expression levels of tens of thousands of genes across tens of thousands of cells, producing data with extremely high dimensionality and complex nonlinear structure. 
To analyze such data effectively, it is essential to employ mathematical representations that preserve geometric structure while capturing biologically meaningful variation.

Despite these advances, many widely used analytical approaches, including principal component analysis (PCA)~\cite{bro2014principal}, 
t-distributed stochastic neighbor embedding (t-SNE)~\cite{maaten2008visualizing}, 
and non-negative matrix factorization (NMF)~\cite{lee1999learning}, 
still represent each cell as a relatively high-dimensional vector in Euclidean space and perform dimension reduction or clustering based on Euclidean similarity. 
Although these Euclidean formulations are easy to implement, they cannot fully capture the correlations and structured relationships among cells that define the organization of high-dimensional single-cell populations. 

Graph-based deep learning models, such as graph neural networks (GNNs)~\cite{li2025graph}, attempt to address this limitation by constructing cell–cell graphs and propagating information through message passing, 
yet their representations remain in Euclidean space and thus fail to describe the intrinsic non-Euclidean geometry of single-cell manifolds.

Many single-cell analysis methods also rely on scale-related parameters that determine the resolution with which structural patterns are captured. 
Different scales reveal distinct biological or geometric characteristics: smaller scales emphasize local cellular transitions and fine-grained variability, whereas larger scales capture global organization, lineage relationships, and inter-cluster structure. 
For instance, methods such as UMAP, t-SNE, and topological PCA adjust the   size to control the level of detail in their embeddings.

In recent years, the development of topological and geometric methods for single-cell analysis has received increasing attention \cite{venkat2023multiscale,sritharan2021computing}. 
These approaches aim to model the intrinsic structure and dynamics of cellular populations through mathematically grounded principles, offering more interpretable and theoretically consistent analyses compared to purely empirical algorithms \cite{ren2025interpretability}. 
For example, the Hodge decomposition of single-cell RNA velocity~\cite{su2024hodge} applies discrete exterior calculus to decompose cellular flow fields into curl-free, divergence-free, and harmonic components, providing a topologically interpretable description of cell-state transitions. 
Similarly, the topological principal component analysis (tPCA) framework~\cite{cottrell2024k} integrates persistent Laplacian operators with spectral regularization to capture multiscale geometric relationships in transcriptomic manifolds. 
From a topological perspective, scGeom~\cite{huynh2024topological} leverages curvature and persistent homology to reveal higher-order geometric organization among cells, 
while topological nonnegative matrix factorization (tNMF)~\cite{hozumi2024analyzing} extends classical NMF by incorporating persistent homology for topologically informed feature extraction. 
In addition, the geometric structure–guided NMF model~\cite{chen2022geometric} combines matrix factorization with manifold regularization to improve the geometric interpretability of gene deconvolution. 
Together, these works highlight the growing importance of topological and geometric representations as mathematical tools for uncovering the complex organization of single-cell systems.

Other mathematical frameworks, including multiscale clustering approaches (MCIST) \cite{cottrell2025multiscale} and optimal transport \cite{cang2023screening} were proposed for spatial transcriptomic data analysis. Both MCIST and graph neural networks \cite{li2025graph}) achieve similar effects by varying the connectivity or depth in the underlying graph. Even in CCP-based methods, the number of gene clusters can be tuned to obtain multiscale features that reflect different biological resolutions. Recent multiscale and multi-fusion frameworks, such as the single-cell multiscale clustering framework (scMSCF)~\cite{jiang2025robust} and multi-fusion graph neural network (scMFGNN)~\cite{yang2025towards}, attempt to integrate information from different scales or omics layers.  However, such concatenation-based strategies often treat multi-view features as independent and ignore their geometric relationships, limiting their ability to represent the hierarchical and correlated structures of single-cell data. These observations suggest that developing new representations of single-cell data in non-Euclidean spaces is a promising direction for capturing multiscale and multi-view biological information more effectively.

 Grassmann manifolds provide a natural and general mathematical framework for representing structured and correlated data. 
Unlike Euclidean representations that treat each feature as an independent coordinate, Grassmann manifolds represent data as linear subspaces that inherently capture correlations and shared structures among multiple views. 
Formally, the Grassmann manifold, denoted as $\Gr(p,n)$, is the set of all $p$-dimensional linear subspaces of $\mathbb{R}^n$, where each subspace corresponds to a single point on the manifold. 
This subspace-based formulation provides a coherent geometric representation for data exhibiting structured dependencies~\cite{L99}. 
Grassmann manifolds have been successfully applied in computer vision and pattern recognition~\cite{chakraborty2015recursive}, 
where subspace representations capture intrinsic variations such as pose, illumination, or temporal dynamics. 
More recently, Li et al.~\cite{li2024exploring} have extended Grassmann manifold geometry to the biological domain by representing genome sequences as points on a Grassmann manifold derived from frequency chaos game representations (FCGRs), providing a geometric perspective for comparing genomic structures. 

However, the application of Grassmann manifolds to single-cell data analysis remains largely unexplored. Unlike images or genomic sequences, there is no straightforward or biologically intuitive way to represent an individual cell as a subspace, which poses a challenge in adapting this framework to single-cell applications. 
Nevertheless, the ability of Grassmann manifolds to capture multi-view correlations and encode structured variability across scales makes them a compelling mathematical foundation for representing single-cell data. 
Results obtained at a single scale are often insufficient to capture the full structural complexity of single-cell data, as each scale provides only a partial view of the underlying manifold. 
To address this limitation, we propose a general framework based on the Grassmann manifold that aggregates representations obtained at multiple scales into a unified subspace. 
This design integrates complementary geometric information across scales, providing a more comprehensive and stable representation of single-cell data.

In this work, we propose a new multiscale Grassmann manifolds framework for representing single-cell data. Each cell is modeled as a point on the Grassmann manifold constructed from its multiscale embeddings. 
The framework introduces a scale sampling function that selects and combines   scales to balance computational efficiency and information richness. 
Pairwise similarities between cells are then computed using Grassmann manifold metrics such as the geodesic or chordal distance, and clustering is performed directly within the Grassmann manifold. Through experiments on several benchmark single-cell RNA-seq datasets, we demonstrate that the proposed multiscale Grassmann manifolds framework produces stable embeddings, improves clustering performance, and provides a more coherent biological organization of cells.

The remainder of this paper is organized as follows.  In Section~2, we present the mathematical foundations underlying this work, including the definitions of the Grassmann manifold, its geometric structures and distance metrics, and the concept of subspace representation. 
Section~3 then describes the proposed framework for representing single-cell data in the Grassmann manifold and explains how multiscale subspaces are constructed through a scale sampling function. 
Next, Section~4 presents experimental results on benchmark single-cell datasets and compares the performance of the proposed multiscale Grassmann manifolds framework with several existing methods. 
A detailed discussion of the framework is provided in Section~5, 
and the paper concludes with final remarks in Section~6.

The main contributions of this paper are summarized as follows:
\begin{enumerate}
   \item We propose the multiscale Grassmann manifolds framework for representing single-cell data, which integrates embeddings from multiple   scales into a unified subspace representation.
   \item We design a power-based scale sampling function that adaptively selects   sizes to balance local and global geometric information across scales.
   \item The proposed framework is flexible and extendable, and can be combined with alternative dimensionality reduction methods capable of capturing multi-view information via the varying of   scales. 
\end{enumerate}

% *** NOTE: Major revision required for this subsection. Clarify definitions and improve mathematical formulation. ***
\section{Mathematical Foundations}

This section presents the mathematical foundations of the proposed framework. 
We first introduce the definition and equivalent formulations of the Grassmann manifold, which provide the geometric setting for our model. 
Next, we describe its key geometric structures, including the tangent space, Riemannian metric, and geodesic formulation, followed by commonly used distance metrics that quantify similarities between subspaces. 
Finally, we explain how data can be represented as subspaces on the Grassmann manifold, linking its geometry to multiscale data modeling. 
The main notation used throughout this paper is summarized in Table~\ref{tab:notation}.

 \begin{table}[h]
   \centering
   \begin{tabular}{c|l}
      \toprule
      \textbf{Symbol}   & \textbf{Description} \\ 
      \midrule
      $\mathbb{R}$ & Real number field \\
      $\mathbb{R}^n$ & $n$-dimensional real vector space \\
      $\mathbb{R}^{n\times n}$ & Set of all real $n\times n$ matrices \\
      $A^T$ & Transpose of matrix $A$ \\
      $\Gr(n,p)$ & Grassmann manifold of all $p$-dimensional subspaces in $\mathbb{R}^n$ \\
       $\Theta(\mathcal{X}, \mathcal{Y})$ & Principal angles between two subspaces $\mathcal{X}$ and $\mathcal{Y}$ \\
      $\dim(\mathcal{U})$ & Dimension of subspace $\mathcal{U}$ \\
      $\rank(P)$ & Rank of matrix $P$ \\
      %$\|\cdot\|_2$ & L$2$-norm \\
      $\operatorname{span}(\cdot)$ & Linear span of the given vectors \\
      $ T_P\Gr(n,p) $ & Tangent space at $p\in \Gr(n,p)$ \\
      $\mathrm{O}(p)$ & Orthogonal group of order $p$ \\
      $\mathrm{GL}(n)$ & General linear group of order $n$ \\
      $\gamma(t)$ & Geodesic curve on $\Gr(n,p)$ parameterized by $t$ \\
      $\dot{\gamma}(t)$ & Tangent (velocity) vector of $\gamma(t)$ at parameter $t$ \\
      $I_p$ & $p\times p$ identity matrix \\
      $\operatorname{Tr}(\cdot)$ & Trace operator of a square matrix \\
      $\exp(\cdot)$ & Matrix exponential map \\
      $\log(\cdot)$ & Principal matrix logarithm \\
      %$d_{\mathrm{chord}}$, $d_{\mathrm{geo}}$ & Chordal and geodesic distances on $\Gr(n,p)$ \\
      $\Phi$ & Subspace embedding map $\Phi:\mathfrak{X}\mapsto P \in \Gr(n,p)$ \\
     % \operatorname{MDR}(X; s_i)$ &   a  multiscale dimension reduction (MDR) operation applied to the single-cell dataset $X$ under scale parameter $s_i$ \\
      \bottomrule
   \end{tabular}
   \caption{Notations used in this paper.}
   \label{tab:notation}
\end{table}

\subsection{Definitions of the Grassmann Manifold}
\label{subsec:def_grassmann}

The Grassmann manifold can be characterized in several mathematically equivalent ways,
each describing its elements as $p$-dimensional linear subspaces of $\mathbb{R}^n$~\cite{BT2024, AMS04, BK2015}. 
Formally, each element of $\Gr(n,p)$ represents a $p$-dimensional linear subspace of $\mathbb{R}^n$, defined as
\[
\Gr(n,p) = \{\, \mathcal{U} \subset \mathbb{R}^n \mid \dim(\mathcal{U}) = p \,\}.
\]
This provides an abstract geometric definition of the manifold as the collection of all $p$-dimensional subspaces. 
The Grassmann manifold admits several equivalent perspectives for describing its elements as $p$-dimensional subspaces of $\mathbb{R}^n$.
These perspectives emphasize different aspects of the same geometric object and are often used interchangeably in both theoretical analysis and applications.

\noindent\textbf{(1) Basis perspective.}  
A subspace $\mathcal{U}\in\Gr(n,p)$ can be represented by any full-rank matrix $Y\in\mathbb{R}^{n\times p}$ whose columns form a basis of $\mathcal{U}$.  
Two matrices $Y_1$ and $Y_2$ define the same subspace if there exists a nonsingular matrix $R\in\mathrm{GL}(p)$ such that $Y_1=Y_2R$, meaning that all matrices whose columns span the same subspace are equivalent under right multiplication by an invertible transformation.  
This perspective identifies $\Gr(n,p)$ with the equivalence class of all rank-$p$ matrices whose column spaces coincide.  
(See, e.g.,~\cite{AMS04} for a general overview.)

\noindent\textbf{(2) Orthonormal basis (ONB) perspective.}  
Alternatively, each subspace can be described by a matrix $U\in\mathbb{R}^{n\times p}$ whose columns form an orthonormal basis of $\mathcal{U}$, satisfying $U^\top U = I_p$.  
All matrices related by right multiplication with an orthogonal matrix $R\in\mathrm{O}(p)$ represent the same subspace, i.e., $UR$ and $U$ correspond to the same point on $\Gr(n,p)$.  
This orthonormal form is particularly convenient for analysis involving differential geometry or optimization on manifolds~\cite{edelman1998geometry}.

\noindent\textbf{(3) Projector perspective.}  
Each subspace $\mathcal{U}$ can also be uniquely characterized by its orthogonal projection operator $P:\mathbb{R}^n\to\mathbb{R}^n$ onto $\mathcal{U}$,  
given by
\[
P = UU^\top,
\]
where $U$ is any orthonormal basis of $\mathcal{U}$.  
The collection of all such projectors forms an equivalent representation of the Grassmann manifold:
\[
\Gr(n,p) = \{\, P\in\mathbb{R}^{n\times n} \mid P^2=P,\; P^\top=P,\; \operatorname{rank}(P)=p \,\}.
\]
This formulation embeds $\Gr(n,p)$ into the space of symmetric matrices and serves as a foundation for defining distances and inner products between subspaces~\cite{BK2015}.

\vspace{0.5em}
While these perspectives differ in representation, they describe the same geometric structure of the Grassmann manifold.
Each viewpoint serves different computational purposes: the basis form emphasizes linear independence, the ONB form ensures orthogonality and numerical stability, and the projector form conveniently expresses geometric relations between subspaces.

\subsection{Geometric Structure and Distance Metrics}\label{subsec:geom}
In this section, we review key geometric structures of the Grassmann manifold based on its projector representation, which is particularly suitable for expressing its differential and Riemannian properties in matrix form
\[
\Gr(n,p) = \{\, P \in \mathbb{R}^{n\times n} \mid P^2 = P,\; P^\top = P,\; \operatorname{rank}(P) = p \,\}.
\]
Each point $P$ represents an orthogonal projector onto a $p$-dimensional subspace of $\mathbb{R}^n$.  
This formulation is particularly convenient for expressing differential and geometric properties in matrix form.

\noindent{\textbf{Tangent space}.}
At a point $P \in \Gr(n,p)$, the tangent space is given by
\begin{equation}\label{eq:tangent}
T_P\Gr(n,p) = \{\, [\Omega, P] = \Omega P - P\Omega \mid \Omega^\top = -\Omega \,\},
\end{equation}
where $\Omega$ belongs to the set of skew-Hermitian matrices.  
Equivalently, every tangent vector $\Delta \in T_P\Gr(n,p)$ can be written in block form
\[
\Delta = Q
\begin{bmatrix}
0 & B^\top \\[0.3em]
B & 0
\end{bmatrix}
Q^\top, 
\qquad B \in \mathbb{R}^{(n-p)\times p},\; Q \in O(n).
\]
The canonical Riemannian metric on $\Gr(n,p)$ is induced from the ambient Euclidean space of matrices.  
For two tangent vectors $\Delta_1 = [\Omega_1, P]$ and $\Delta_2 = [\Omega_2, P]$,  
their inner product is defined by
\begin{equation}\label{eq:metric}
\langle \Delta_1, \Delta_2 \rangle_P
= \tfrac{1}{2}\operatorname{Tr}(\Delta_1^\top \Delta_2)
= \operatorname{Tr}(\Omega_1^\top \Omega_2),
\end{equation}
where $\operatorname{Tr}(\cdot)$ denotes the matrix trace.  
This metric is invariant under the orthogonal group action 
$P \mapsto QPQ^\top$ for any $Q \in O(n)$, ensuring that it depends only on the subspaces, not on the particular matrix representation.  
It provides the geometric foundation for defining geodesics and curvature on $\Gr(n,p)$.

\noindent
\textbf{Geodesic.} 
With the canonical metric in (2), the Grassmann manifold $\Gr(n,p)$ admits a smooth geodesic flow generated by the matrix exponential.
For a point $P \in \Gr(n,p)$ and a tangent direction $\Delta = [\Omega, P] \in T_P\Gr(n,p)$,  
the geodesic $\gamma(t)$ starting at $P$ with initial velocity $\Delta$ is given by \cite{BT2024}
\begin{equation}\label{eq:geodesic_exp}
\gamma(t) = \exp(t\Omega)\, P\, \exp(-t\Omega),
\end{equation}
where $\exp(\cdot)$ denotes the matrix exponential map, and the curve satisfies
\[
\gamma(0) = P, \qquad \dot{\gamma}(0) = \Delta.
\]

\noindent
When the matrix $(I - 2Q)(I - 2P)$ has no negative real eigenvalues,  
the minimizing geodesic between two points $P, Q \in \Gr(n,p)$  
admits the following closed-form expression~\cite{BK2015}:
\begin{equation}\label{eq:geo_closed}
\gamma(t)
= e^{\,\frac{t}{2}\log((I-2Q)(I-2P))}\,
P\,
e^{-\frac{t}{2}\log((I-2Q)(I-2P))},
\qquad t \in [0,1],
\end{equation}
where $\log$ denotes the principal matrix logarithm.  
Equation~\eqref{eq:geo_closed} describes the shortest geodesic from $P$ to $Q$ under the canonical metric whenever such a unique minimizing path exists.

\noindent \textbf{Distance Metrics.}  
The shortest path between two points on $\Gr(n,p)$, defined by the geodesic above, can be equivalently characterized in terms of the relative orientation between subspaces, quantified by a set of principal angles.
\begin{definition}[Principal Angles]
Let $\mathcal{X}, \mathcal{Y} \subset \mathbb{R}^n$ be two subspaces with 
$\dim(\mathcal{X}) = p$ and $\dim(\mathcal{Y}) = q$, and let $m = \min(p,q)$. 
The principal angles between $\mathcal{X}$ and $\mathcal{Y}$ are defined as
\[
\Theta(\mathcal{X}, \mathcal{Y}) = [\theta_1, \ldots, \theta_m], \qquad 
\theta_i \in [0, \tfrac{\pi}{2}], \quad i = 1, \ldots, m,
\]
where $\cos(\theta_i)$ are the singular values of $U_{\mathcal{X}}^\top U_{\mathcal{Y}}$, 
and $U_{\mathcal{X}}, U_{\mathcal{Y}}$ are orthonormal basis matrices for $\mathcal{X}$ and $\mathcal{Y}$, respectively. 
The corresponding vectors $\{x_i\}$ and $\{y_i\}$ that achieve these angles are called the principal vectors.
\end{definition}

The principal angles provide a direct way to quantify the smallest angular deviations between subspaces 
and form the basis for defining various distance measures on $\Gr(n,p)$.  
Different distance functions have been proposed in the literature, 
which can be equivalently expressed using either principal angles or orthonormal basis representations  
\cite{ye2016schubert}.  
Several representative metrics are summarized in Table~\ref{tab:grassmann_distances}.

\begin{table}[!htbp]
\centering
\caption{Common distance metrics on $\Gr(n,p)$ expressed in terms of principal angles 
$\{\theta_i\}_{i=1}^p$ between  $\mathcal{X}$ and $\mathcal{Y}$.}
\label{tab:grassmann_distances}
\renewcommand{\arraystretch}{1.2}
\begin{tabular}{ccc}
\hline
\textbf{Metric} & \textbf{Notation} & \textbf{Definition (principal-angle form)} \\
\hline
Geodesic        & $d_{\mathrm{geo}}(\mathcal{X},\mathcal{Y})$ 
& $\displaystyle \left( \sum_{i=1}^{p} \theta_i^2 \right)^{1/2}$ \\[0.3em]
Chordal         & $d_{\mathrm{chord}}(\mathcal{X},\mathcal{Y})$ 
& $\displaystyle \left( \sum_{i=1}^{p} \sin^2 \theta_i \right)^{1/2}$ \\[0.3em]
Fubini--Study   & $d_{\mathrm{FS}}(\mathcal{X},\mathcal{Y})$ 
& $\displaystyle \cos^{-1}\!\left( \prod_{i=1}^{p} \cos \theta_i \right)$ \\[0.3em]
Martin          & $d_{\mathrm{Martin}}(\mathcal{X},\mathcal{Y})$ 
& $\displaystyle \left( \log \prod_{i=1}^{p} \frac{1}{\cos^2 \theta_i} \right)^{1/2}$ \\[0.3em]
Procrustes      & $d_{\mathrm{Proc}}(\mathcal{X},\mathcal{Y})$ 
& $\displaystyle 2 \left( \sum_{i=1}^{p} \sin^2(\tfrac{\theta_i}{2}) \right)^{1/2}$ \\
\hline
\end{tabular}
\end{table}

%These distance metrics form the foundation for the Grassmann manifold  framework proposed in this study.

\subsection{Subspace Representation}

In Euclidean models, each data instance is represented by a single vector $x_i \in \mathbb{R}^n$, 
and similarity is typically measured using Euclidean or correlation-based distances. 
Such point-wise representations, however, neglect internal structure, correlations, and multi-view variability that may exist within a sample or across multiple observations. 
A more general and geometrically consistent approach is to represent each instance by a low-dimensional subspace, 
which lies on the Grassmann manifold $\Gr(n,p)$.

\begin{definition}[Subspace Representation]
Let $\mathfrak{X} = \{x_1, x_2, \ldots, x_m\} \subset \mathbb{R}^n$ be a set of feature vectors associated with a single object, cell, or sample measured under different conditions. 
A \emph{subspace representation} of $\mathfrak{X}$ is defined as the linear subspace
\[
\mathcal{S} = \mathrm{span}(Y), \qquad Y = [x_1, x_2, \ldots, x_p] \in \mathbb{R}^{n\times p}, \; \mathrm{rank}(Y)=p.
\]
After orthonormalization, the basis matrix $\tilde{Y}$ satisfies $\tilde{Y}^\top \tilde{Y} = I_p$, 
and the subspace $\mathcal{S}$ corresponds to a point on the Grassmann manifold $\Gr(n,p)$.
\end{definition}

Equivalently, a subspace can be represented by its orthogonal projection matrix
\[
P = \tilde{Y}\tilde{Y}^\top, \qquad P^\top = P, \; P^2 = P, \; \operatorname{rank}(P) = p,
\]
which uniquely characterizes $\mathcal{S}$ and enables the use of  distance metrics in Table \ref{tab:grassmann_distances} for comparing different samples. 
The mapping
\[
\Phi : \mathfrak{X} \longmapsto P \in \Gr(n,p)
\]
is referred to as the \emph{subspace embedding map}, 
which projects a collection of feature vectors into the space of $p$-dimensional subspaces.

Subspace representations possess several useful mathematical properties. 
They are invariant under orthogonal transformations of the basis, 
i.e., $\mathrm{span}(Y) = \mathrm{span}(YR)$ for any $R \in \mathrm{O}(p)$, 
and they encode higher-order correlations among the columns of $Y$, 
providing a compact description of structured or multi-view data. 
When distances between subspaces are measured using metrics on $\Gr(n,p)$, 
the resulting geometry captures relationships that are robust to noise, rotation, and small perturbations in the original feature space.

\section{Multiscale Representation of Single-Cell Data on the Grassmann Manifold}

This section introduces the proposed multiscale Grassmann manifolds (MGM) framework, 
which provides a unified geometric representation of single-cell data by integrating information obtained from multiple scales. 
We first present the overall procedure for representing single-cell data as points on the Grassmann manifold through embeddings constructed under multiple representation scales. 
An overview of the workflow is illustrated in Figure~\ref{fig:framework}. 
Next, we describe a scale sampling function that determines the set of scales used to generate multiscale embeddings. 
Finally, the resulting Grassmann manifold representations yield a distance matrix that can be directly applied to downstream analyses, 
including clustering or classification, by replacing Euclidean metrics with Grassmann manifold distances.

\subsection{Framework Overview of Multiscale Grassmann Manifolds Framework for Single-Cell
Representation}

Given a single-cell expression matrix 
\[
X = 
\begin{pmatrix}
x_1\\
x_2\\
\vdots\\
x_M
\end{pmatrix}
\in \mathbb{R}^{M\times N},
\]
where $M$ denotes the number of cells, $N$ denotes the number of genes, and each row vector $x_j \in \mathbb{R}^{1\times N}$ represents the gene expression profile of cell $j$ for $j=1,2,\ldots,M$, 
the goal of the framework is to construct a Grassmann manifold-based representation that captures the intrinsic structure of the dataset across multiple  scales.
\vspace{0.5em}

\noindent\textbf{Stage 1: Multiscale embeddings.}  
The first step of the multiscale Grassmann manifolds framework is to generate low-dimensional representations that capture cellular features under multiple   or resolution scales. 
Let $p$ denote the number of scales considered in the multiscale setting, that is, the total number of embeddings generated under different scale parameters. 
Each scale produces an embedding of dimension $n$, which represents the number of features preserved for each cell after applying a suitable dimension reduction method. 
The set of  scales is defined as 
\[
S = \{s_1, s_2, \ldots, s_p\},
\]
where each $s_i$ is determined by a sampling function $f_{\text{sample}}$, which will be introduced in Section~3.2. 
For all scales, the embedding dimension $n$ is fixed to ensure consistency across representations.  
Applying the selected dimension reduction method under the scales $s_1, s_2, \ldots, s_p$ yields a sequence of feature matrices
\[
E_i =
\begin{pmatrix}
z^{(i)}_1\\
z^{(i)}_2\\
\vdots\\
z^{(i)}_M
\end{pmatrix}
= \operatorname{MDR}(X; s_i) \in \mathbb{R}^{M\times n}, 
\quad i = 1, 2, \ldots, p,
\]
where $\operatorname{MDR}(X; s_i)$ denotes a  multiscale dimension reduction (MDR) operation applied to the single-cell dataset $X$ under scale parameter $s_i$.  
Each row vector $z^{(i)}_j \in \mathbb{R}^{1\times n}$ represents the embedding of cell $j$ at scale $s_i$.  
Each matrix $E_i$ therefore contains the $n$-dimensional embeddings of all $M$ cells under the corresponding scale.  
These multiscale embeddings provide complementary geometric views of the dataset, 
where smaller scales emphasize fine-grained local transitions, 
while larger scales capture broader global organization across the cellular manifold.

\vspace{0.5em}

\noindent\textbf{Stage 2: Feature aggregation.}  
For each cell $j = 1, 2, \ldots, M$, we collect its feature vectors obtained from all $p$ embeddings generated under different scales. 
These multiscale features are concatenated column-wise to form a matrix
\[
Z_j =
\begin{bmatrix}
(z^{(1)}_j)^\top & (z^{(2)}_j)^\top & \cdots & (z^{(p)}_j)^\top
\end{bmatrix} 
\in \mathbb{R}^{n\times p},
\]
where $z^{(i)}_j \in \mathbb{R}^{1\times n}$ denotes the embedding vector of cell $j$ obtained at scale $s_i$.  
The matrix $Z_j$ integrates multiple geometric views of the same cell across different scales, 
thereby capturing complementary information that would otherwise be lost in a single-scale representation.

\vspace{0.5em}
\noindent\textbf{Stage 3: Grassmann manifold representation.}  
After assembling the multiscale feature matrix $Z_j$ for each cell, we interpret its column space as a subspace in $\mathbb{R}^n$.  
Formally, the span of the feature vectors across all scales defines
\[
G_j = \operatorname{span}(Z_j) = 
\operatorname{span}\{\, (z^{(1)}_j)^\top, (z^{(2)}_j)^\top, \ldots, (z^{(p)}_j)^\top \,\} \in \Gr(n,p),
\]
where each $z^{(i)}_j$ corresponds to the embedding of cell $j$ obtained at scale $s_i$.  
The resulting subspace $G_j$ characterizes cell $j$ in a multiscale feature space, 
and all cells are thus represented as points on the Grassmann Manifold $\Gr(n,p)$. 

\vspace{0.5em}
\noindent\textbf{Stage 4: Distance computation on the Grassmann manifold.}  
Based on the subspace representation $\{G_j\}_{j=1}^M$, we compute pairwise distances between all cells using the Grassmann manifold metrics introduced in Section~2. 
Commonly used distance measures include the geodesic distance and the chordal distance, 
both of which quantify the geometric discrepancy between two subspaces in $\Gr(n,p)$.  
For each pair of cells $(i,j)$, the Grassmann manifold distance $d(G_i, G_j)$ is computed according to the selected metric in Table \ref{tab:grassmann_distances}, 
forming a symmetric distance matrix
\[
D = [d_{ij}]_{i,j=1}^M=[\, d(G_i, G_j) \,]_{i,j=1}^M \in \mathbb{R}^{M\times M}.
\]
This distance matrix serves as the foundation for downstream analyses, 
such as clustering, visualization, or classification, 
by providing a non-Euclidean measure of similarity that reflects the multiscale geometric relationships between cells.

\begin{figure}[ht]
    \centering
    \hspace*{0cm}
    \includegraphics[width=0.9\textwidth]{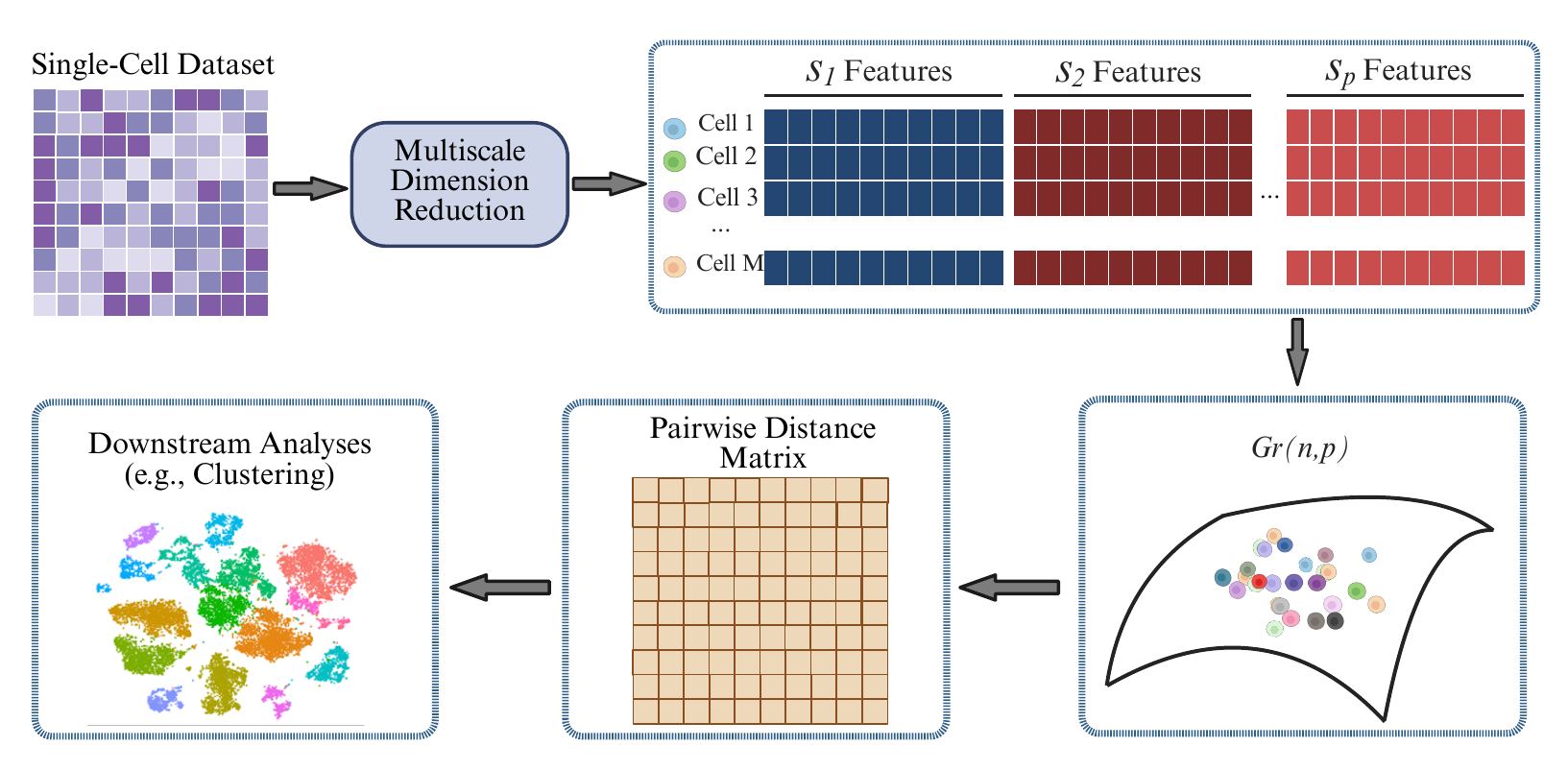}
   \caption{Overview of the proposed multiscale Grassmann manifolds (MGM) framework. 
Starting from a single-cell gene expression matrix, 
multiple low-dimensional embeddings are generated under different neighborhood sizes ($s_1,\ldots,s_p$) using a chosen dimensionality reduction method. 
For each cell, the resulting multiscale feature vectors are aggregated into a matrix whose column space defines a subspace on the Grassmann manifold $\Gr(n,p)$. 
Pairwise distances between these subspaces are then computed to form a pairwise distance matrix, 
which can be used for downstream analyses such as clustering.}
\label{fig:framework}
\end{figure}

\noindent
Algorithm~\ref{alg:mgm_repr} summarizes the complete multiscale Grassmann manifolds (MGM) representation procedure, 
and the overall workflow is illustrated in Figure~\ref{fig:framework}.

\begin{algorithm}
\caption{Multiscale Grassmann Manifolds  (MGM)}
\label{alg:mgm_repr}
\hrule\vspace{0.3em}
\begin{algorithmic}[1]
    \State \textbf{Input:} Single-cell data matrix $X \in \mathbb{R}^{M\times N}$; 
     scale set $S=\{s_1, s_2, \ldots, s_p\}$; embedding dimension $n$
    \State \textbf{Output:} Cell representations $\{G_j\}_{j=1}^M \subset \Gr(n,p)$ and pairwise distance matrix $D\in\mathbb{R}^{M\times M}$

    \Statex
    \textbf{Stage 1: Multiscale dimension reduction}
    \For{$i = 1$ to $p$}
        \State $E_i \gets \operatorname{MDR}(X; s_i) \in \mathbb{R}^{M\times n}$ 
        \hfill \(\triangleright\) $E_i = [z^{(i)}_1; z^{(i)}_2; \ldots; z^{(i)}_M]$
    \EndFor

    \Statex
    \textbf{Stage 2: Feature aggregation}
    \For{$j = 1$ to $M$}
        \State $Z_j \gets \big[(z^{(1)}_j)^{\top}, (z^{(2)}_j)^{\top}, \ldots, (z^{(p)}_j)^{\top}\big] \in \mathbb{R}^{n\times p}$
        \hfill \(\triangleright\) multiscale feature matrix for cell $j$
    \EndFor

    \Statex
    \textbf{Stage 3: Grassmann manifold representation}
    \For{$j = 1$ to $M$}
        \State $G_j \gets \operatorname{span}(Z_j)$ 
        \hfill \(\triangleright\) interpret each $Z_j$ as a $p$-dimensional subspace on $\Gr(n,p)$
    \EndFor

    \Statex
    \textbf{Stage 4: Distance computation}
    \State Compute pairwise distances 
    $d_{jk} \gets d_{\Gr}(G_j, G_k)$ 
    using Grassmann Manifold metrics 
    (e.g., geodesic or chordal distance).

    \State \Return $\{G_j\}_{j=1}^M$, $D$
\end{algorithmic}
\vspace{0.3em}\hrule
\end{algorithm}

\subsection{Sampling Function for Multiscale Selection}

To generate the  scales used by the framework, we adopt a power-based sampling function that flexibly controls the density of sampled scales across a prescribed interval. 
Given a minimum and maximum   size $a=s_{\min}$ and $b=s_{\max}$, a target number of samples $n_{\text{scales}}\!\ge 2$, and a fixed exponent $p_s>0$, we first compute raw (real-valued) samples
\begin{equation}
\label{eq:power-sampling}
\tilde s_i \;=\; a + (b-a)\,t_i^{\,p_s}, 
\qquad 
t_i \;=\; \frac{i}{\,n_{\text{scales}}-1\,}, 
\quad i=0,1,\ldots,n_{\text{scales}}-1 .
\end{equation}
We then quantize $\{\tilde s_i\}$ to integer   sizes and remove duplicates to obtain the final, ordered scale set
\[
S \;=\; \{\, s_1 < s_2 < \cdots < s_p \,\} \subset \{a,a+1,\ldots,b\}.
\]
Here $p=\lvert S\rvert$ is the number of distinct scales returned by the sampling function, which may satisfy $p \le n_{\text{scales}}$ due to quantization and de-duplication. 
Throughout the framework, $p$ also determines the number of multiscale views per cell and hence the rank on $\Gr(n,p)$. 
To avoid notational conflict, $p_s$ denotes the sampling exponent and should not be confused with $p$. 
In practice we typically choose $p_s>1$ (e.g., $p_s=1.6$) to bias the sampling toward smaller  s (denser local views) while still covering the full range $[a,b]$. Algorithm \ref{alg:power-sampling}  implements the power-based sampling $f_{\text{sample}}$.
\vspace{0.5em}
\setcounter{algorithm}{1}

\begin{algorithm}
\caption{Power-based sampling function $f_{\text{sample}}(a,b,n_{\text{scales}},p_s)$}
\label{alg:power-sampling}
\hrule\vspace{0.3em}
\begin{algorithmic}[1]
    \State \textbf{Input:} bounds $a,b\in\mathbb{N}$ with $2\le a<b$; number of samples $n_{\text{scales}}\!\ge 2$; exponent $p_s>0$
    \State \textbf{Output:} ordered, unique scales $S=\{s_1,\ldots,s_p\}$ with $a\le s_1<\cdots<s_p\le b$
    \State Generate $t_i \gets i/(n_{\text{scales}}-1)$ for $i=0,1,\ldots,n_{\text{scales}}-1$
    \State Compute raw values $\tilde s_i \gets a + (b-a)\, t_i^{\,p_s}$
    \State Quantize $q_i \gets \mathrm{round}(\tilde s_i)$ and clip: $q_i \gets \min\{\max\{q_i,a\},\,b\}$
    \State Form $S \gets$ the sorted set of unique values in $\{q_i\}_{i=0}^{n_{\text{scales}}-1}$ \hfill \(\triangleright\) now $p=\lvert S\rvert \le n_{\text{scales}}\)
    \State \Return $S=\{s_1,\ldots,s_p\}$
\end{algorithmic}
\vspace{0.3em}\hrule
\end{algorithm}
\vspace{0.5em}
\begin{remark}
The exponent $p_s$ in the power-based sampling function controls the density of sampled scales across the range $[a,b]$.  
When $p_s = 1$, the sampling is linear, producing uniformly spaced scales.  
For $p_s < 1$, more scales are concentrated toward the larger   sizes (favoring global structure), 
whereas for $p_s > 1$, the sampling becomes denser near smaller  s (emphasizing local information).  
In our experiments, for datasets with a relatively small number of cells (for datasets with fewer than or around 1000 cells), 
we set $a = 5$, $b = n$, and use $p_s = 1.6 > 1$ to obtain finer resolution among smaller-scale  s while maintaining coverage of the full range.
\end{remark}

\subsection{Clustering and Performance Evaluation}

Once each cell is represented as a point on the Grassmann manifold $\Gr(n,p)$, 
the resulting pairwise distance matrix provides a flexible foundation 
for various downstream analyses. 
Standard clustering or classification algorithms can be readily extended 
by substituting the Euclidean distance with a Grassmann manifold metric. 
This allows clustering to be performed directly in the manifold space, 
where the subspace geometry captures intrinsic relationships among samples 
more effectively than Euclidean representations.

A number of clustering and recognition methods based on the Grassmann manifold geometry 
have been proposed in related fields. 
One of the earliest examples is the Grassmann Nearest Neighbor (GNN) algorithm 
introduced by Yamaguchi et~al.~\cite{yamaguchi1998face}, 
which classifies each point on the manifold by comparing its geodesic distance 
to the nearest class subspace. 
Subsequent studies extended this concept to unsupervised settings, 
including $k$-means clustering on Grassmann manifolds~\cite{li2016discriminative} 
and spectral clustering based on Grassmann manifold affinity matrices~\cite{kumar2011co, yang2025multi}. 
These methods leverage the manifold’s intrinsic geometry 
to improve robustness to noise and nonlinear variation.

To evaluate clustering performance,  
we employ five commonly used metrics in single-cell analysis:  
clustering accuracy (ACC)~\cite{crouse2016implementing},  
normalized mutual information (NMI)~\cite{vinh2009information},  
adjusted Rand index (ARI)~\cite{hubert1985comparing},  
purity~\cite{rao2018exploring}, and average purity (Avg-Purity)~\cite{suwayyid2025cakl}.  
The definitions of these metrics are provided in the Supporting Information.  
Each metric compares the predicted cluster labels with the ground-truth cell labels, 
offering complementary perspectives on clustering accuracy, label consistency, and partition quality.  
Detailed results and comparisons with baseline methods  
are presented in Section~4.

\section{Experiments}

This section presents the experimental evaluation of the proposed multiscale Grassmann manifolds (MGM) framework 
and its performance on benchmark single-cell RNA-seq datasets.  
We aim to examine how the manifold-based representation improves single-cell data analysis 
by comparing its clustering performance with standard Euclidean and manifold-based approaches.  
All experiments were conducted on publicly available benchmark datasets, 
and comparative analyses were performed under consistent preprocessing and clustering conditions.

\subsection{Datasets and Experimental Setup}

The proposed framework was evaluated on nine publicly available human single-cell RNA-seq datasets 
that span a range of sizes—from small datasets with fewer than 50 cells 
to large datasets with nearly 2,000 cells.  
These datasets have been widely used in previous studies 
(e.g.,~\cite{feng2024multiscale, hozumi2024analyzing}) 
and are considered reliable benchmarks for evaluating dimensionality reduction and clustering performance.  
 clustering performance.

Table~\ref{tab:datasets} summarizes the main characteristics of the datasets, 
including the number of cells and genes.  
This collection provides a balanced test bed for evaluating the method across different data scales.

\begin{table}[!htbp]
\centering
\caption{Summary of benchmark single-cell RNA-seq datasets used in the experiments. 
Each dataset includes its GEO accession number, reference, number of annotated cell types, 
and the numbers of samples (cells) and genes.}
\label{tab:datasets}
\begin{tabular}{lcccc}
\hline
\textbf{GEO accession} & \textbf{Reference} & \textbf{Cell types} & \textbf{Samples} & \textbf{Genes} \\
\hline
GSE75748time & Chu~\cite{chu2016single} & 6 & 758 & 19\,189 \\
GSE94820 & Villani~\cite{villani2017single} & 5 & 1\,140 & 26\,593 \\
GSE67835 & Dramanis~\cite{darmanis2015survey} & 8 & 420 & 22\,084 \\
GSE75748cell & Chu~\cite{chu2016single} & 6 & 1\,018 & 19\,189 \\
GSE109979 & Lu~\cite{lu2018single} & 4 & 329 & 18\,840 \\
GSE84133human1 & Baron~\cite{baron2016single} & 9 & 1\,895 & 20\,125 \\
GSE84133human2 & Baron~\cite{baron2016single} & 9 & 1\,702 & 20\,125 \\
GSE84133human4 & Baron~\cite{baron2016single} & 6 & 1\,275 & 20\,125 \\
GSE57249 & Biase~\cite{biase2014cell} & 3 & 49 & 25\,737 \\
\hline
\end{tabular}
\end{table}

For each dataset, gene expression matrices were normalized and log-transformed 
according to standard preprocessing procedures.  To systematically evaluate the performance of the proposed framework, 
we designed two experimental configurations that differ in preprocessing rigor and noise levels. 
The first setup (noisy condition) tests the robustness of the model under high-noise data, 
while the second setup (refined condition) focuses on well-preprocessed data 
to examine representational quality in cleaner manifolds. 
In both settings, the proposed method was compared with widely used dimensionality reduction and clustering baselines.

In the noisy setting, the proposed method was compared with  Avg-UMAP 
(obtained by averaging the embeddings across all scales) and PCA. In the refined setting, results for Nonnegative Matrix Factorization (NMF) 
and t-SNE were directly adopted from previous studies~\cite{feng2024multiscale, hozumi2024analyzing} 
for baseline comparison. 
Both methods were evaluated using k-means as the downstream algorithm, 
ensuring consistency with our experimental setup.

 \vspace{0.3em}
\noindent\textbf{Setup I: Noisy condition.}
In this configuration, single-cell expression matrices were first reduced to 200 dimensions by PCA 
and then embedded into a 100-dimensional space using a multiscale dimensionality reduction process.  
Neighborhood scales were generated by the power-based sampling function with exponent $p_s=2$.  
For most datasets, the number of sampled scales was set to $n_{\text{scales}}=25$, 
yielding a subspace rank of $p=23$ and a Grassmann representation in $\Gr(100,23)$.  
For the smallest dataset, GSE57249, which contains only 49 cells, 
we used $n_{\text{scales}}=13$ and $p=12$ to maintain computational stability.  
This setup evaluates the robustness of the proposed framework 
under conditions where residual background noise remains after preprocessing.  
Nine single-cell RNA-seq datasets were tested:  
GSE75748time, GSE94820, GSE67835, GSE75748cell, GSE109979,  
GSE84133human1, GSE84133human2, GSE84133human4, and GSE57249.  
The method was compared with  Avg-UMAP and PCA under identical settings 
to assess its ability to preserve geometric consistency in noisy embeddings.

\vspace{0.3em}
\noindent\textbf{Setup II: Refined condition.}
In this configuration, single-cell data were carefully preprocessed 
to reduce background noise and retain the most informative features.  
For small to medium datasets, including  
GSE75748time, GSE94820, GSE67835, GSE75748cell, and GSE109979,  
the gene expression matrices were reduced to 50 dimensions by PCA 
and further embedded to $n=20$ dimensions.  
Each cell was then represented by a subspace of rank $p=10$, 
forming points on the Grassmann manifold $\Gr(20,10)$.  
For larger datasets, including GSE84133human1, GSE84133human2, and GSE84133human4,  
PCA reduction to 100 dimensions and embedding to $n=50$ were used, 
with a subspace rank of $p=19$, corresponding to $\Gr(50,19)$.  
For the smallest dataset, GSE57249,  
PCA and embedding dimensions were set to 20 and 15, respectively, 
with a subspace rank of $p=8$ on $\Gr(15,8)$.  
This refined configuration evaluates the representational power of the proposed framework 
when the data manifold is well preserved through preprocessing 
and the impact of noise is minimized.

\vspace{0.3em}
\noindent
Both setups employed the power-based sampling strategy introduced in Section~3.2 to generate   scales, with different exponents reflecting their respective scale ranges. 
For Setup~I, which uses a broader range ($[5,100]$), the power exponent was set to $p_s=2.0$ to achieve denser sampling at smaller   sizes. 
For Setup~II, where the scale interval is narrower, $p_s=1.6$ was used instead. 
In all experiments, the chordal distance was adopted for Grassmann manifold affinity computation. 
All models were implemented in Python and evaluated under the same computational environment to ensure fair and reproducible comparison. The full parameter configurations for Setup I and Setup II are provided in Section~1 of the Supporting Information.

\textbf{Remark.} 
The number of usable scales \(p\) is determined by the sampling function 
\(f_{\text{sample}}\), which may produce redundant   sizes after 
quantization. In practice,   \(p\) denotes the number of unique scales obtained after removing duplicates from the generated set.

In the implementation, multiscale low-dimensional embeddings were generated through a multiscale dimension reduction process, 
where UMAP~\cite{becht2019dimensionality} was employed as the underlying dimensionality reduction method.  
These embeddings were then integrated using the proposed multiscale Grassmann manifolds (MGM) framework 
to form Grassmann manifold representations.  
Pairwise distances between cells were computed on the manifold using the chordal distance.  
For the noisy condition (Setup~I), spectral clustering \cite{ng2001spectral} was used as the downstream algorithm,  
while for the refined condition (Setup~II), $k$-means clustering \cite{mcqueen1967some} was adopted 
to ensure consistency with the baseline methods reported in previous studies. The implementation details of UMAP, as well as the clustering settings including the specific configurations for spectral clustering and $k$-means, are provided in the Supporting Information. 
Unless otherwise specified, the chordal distance was employed as the default Grassmann metric.  
Visualization results were also provided to illustrate the behavior of the framework across multiple scales.

 \subsection{Results} % 4.2

This subsection presents the quantitative results of the proposed multiscale Grassmann manifolds (MGM) framework 
under the two experimental configurations described in Section~4.1.  
All methods were evaluated using five random seeds (\(1, 3, 5, 7, 9\)), 
and the reported values correspond to the averages across these runs.  
Clustering performance was assessed using  standard evaluation metrics: 
accuracy (ACC), normalized mutual information (NMI), adjusted Rand index (ARI), 
purity, and average purity (Avg-Purity).
For each dataset, the best-performing results are indicated in \textbf{bold}.

\subsubsection{Results under the noisy condition (Setup I)}
Under the noisy setting, the proposed multiscale Grassmann manifolds (MGM) framework was evaluated using spectral clustering and compared with PCA and  Avg-UMAP.  
Despite the presence of residual noise after preprocessing, MGM achieved superior or comparable clustering results across most datasets.  
Quantitative results for clustering accuracy (ACC), normalized mutual information (NMI), adjusted Rand index (ARI), and purity are summarized in Tables~\ref{tab:acc_noisy}--\ref{tab:purity_noisy}.

\begin{table}[!htbp]
\centering
\caption{ACC under the noisy condition (Setup I). Best in \textbf{bold}.}
\label{tab:acc_noisy}
\begin{tabular}{lccc}
\hline
Dataset & MGM &  Avg-UMAP & PCA \\
\hline
GSE75748time       & \textbf{0.7219} & 0.6599 & 0.4222 \\
GSE94820           & 0.7314 & \textbf{0.7391} & 0.6496 \\
GSE67835           & \textbf{0.6819} & 0.4970 & 0.5581 \\
GSE75748cell       & \textbf{0.8316} & 0.7568 & 0.3984 \\
GSE109979 & \textbf{0.7732} & 0.7009 & 0.3463 \\
GSE84133human1     & \textbf{0.6920} & 0.6599 & 0.5714 \\
GSE84133human2     & \textbf{0.7053} & 0.5972 & 0.4401 \\
GSE84133human4     & \textbf{0.7677} & 0.7323 & 0.5307 \\
GSE57249         & \textbf{0.9102} & 0.8507 & 0.5388 \\
\hline
Average ACC            & \textbf{0.7573} & 0.6882 & 0.4950 \\
\hline
\end{tabular}
\end{table}

\begin{table}[h]
\centering
\caption{NMI under the noisy condition (Setup I). Best in \textbf{bold}.}
\label{tab:nmi_noisy}
\begin{tabular}{lccc}
\hline
Dataset & MGM &  Avg-UMAP & PCA \\
\hline
GSE75748time       & \textbf{0.6508} & 0.6005 & 0.2831 \\
GSE94820           & \textbf{0.6721} & 0.6472 & 0.4916 \\
GSE67835           & \textbf{0.6044} & 0.4520 & 0.4218 \\
GSE75748cell       & \textbf{0.8164} & 0.7750 & 0.4220 \\
GSE109979 & \textbf{0.5590} & 0.4939 & 0.1319 \\
GSE84133human1     & \textbf{0.7382} & 0.7223 & 0.4677 \\
GSE84133human2     & \textbf{0.7399} & 0.6364 & 0.3673 \\
GSE84133human4     & \textbf{0.7571} & 0.7043 & 0.4010 \\
GSE57249           & \textbf{0.7249} & 0.6589 & 0.2589 \\
\hline
Average  NMI          & \textbf{0.6959} & 0.6323 & 0.3606 \\
\hline
\end{tabular}
\end{table}

\begin{table}[!htbp]
\centering
\caption{ARI under the noisy condition (Setup I). Best in \textbf{bold}.}
\label{tab:ari_noisy}
\begin{tabular}{lccc}
\hline
Dataset & MGM &  Avg-UMAP & PCA \\
\hline
GSE75748time       & \textbf{0.5238} & 0.4795 & 0.1620 \\
GSE94820           & \textbf{0.5876} & 0.5546 & 0.3934 \\
GSE67835           & \textbf{0.4904} & 0.2889 & 0.2657 \\
GSE75748cell       & \textbf{0.6999} & 0.6451 & 0.2028 \\
GSE109979 & \textbf{0.5189} & 0.4518 & 0.0495 \\
GSE84133human1     & \textbf{0.5515} & 0.5084 & 0.2982 \\
GSE84133human2     & \textbf{0.5871} & 0.4545 & 0.2235 \\
GSE84133human4     & \textbf{0.6749} & 0.4570 & 0.2651 \\
GSE57249            & \textbf{0.7293} & 0.6449 & 0.0839 \\
\hline
Average ARI           & \textbf{0.5959} & 0.4983 & 0.2160 \\
\hline
\end{tabular}
\end{table}

\begin{table}[h]
\centering
\caption{Purity under the noisy condition (Setup I). Best in \textbf{bold}.}
\label{tab:purity_noisy}
\begin{tabular}{lccc}
\hline
Dataset & MGM &  Avg-UMAP & PCA \\
\hline
GSE75748time       & \textbf{0.7219} & 0.6946 & 0.4844 \\
GSE94820           & \textbf{0.7693} & 0.7556 & 0.6672 \\
GSE67835           & \textbf{0.7629} & 0.6275 & 0.6314 \\
GSE75748cell       & \textbf{0.8316} & 0.7908 & 0.5477 \\
GSE109979 & \textbf{0.7732} & 0.7049 & 0.3963 \\
GSE84133human1     & 0.8434 & \textbf{0.8524} & 0.7404 \\
GSE84133human2     & \textbf{0.9235} & 0.8285 & 0.6741 \\
GSE84133human4     & \textbf{0.8933} & 0.8441 & 0.6504 \\
GSE57249           & \textbf{0.9102} & 0.8592 & 0.5918 \\
\hline
Average    Purity        & \textbf{0.8255} & 0.7731 & 0.5982 \\
\hline
\end{tabular}
\end{table}

\begin{figure}[h]
  \centering
  \includegraphics[width=1\textwidth]{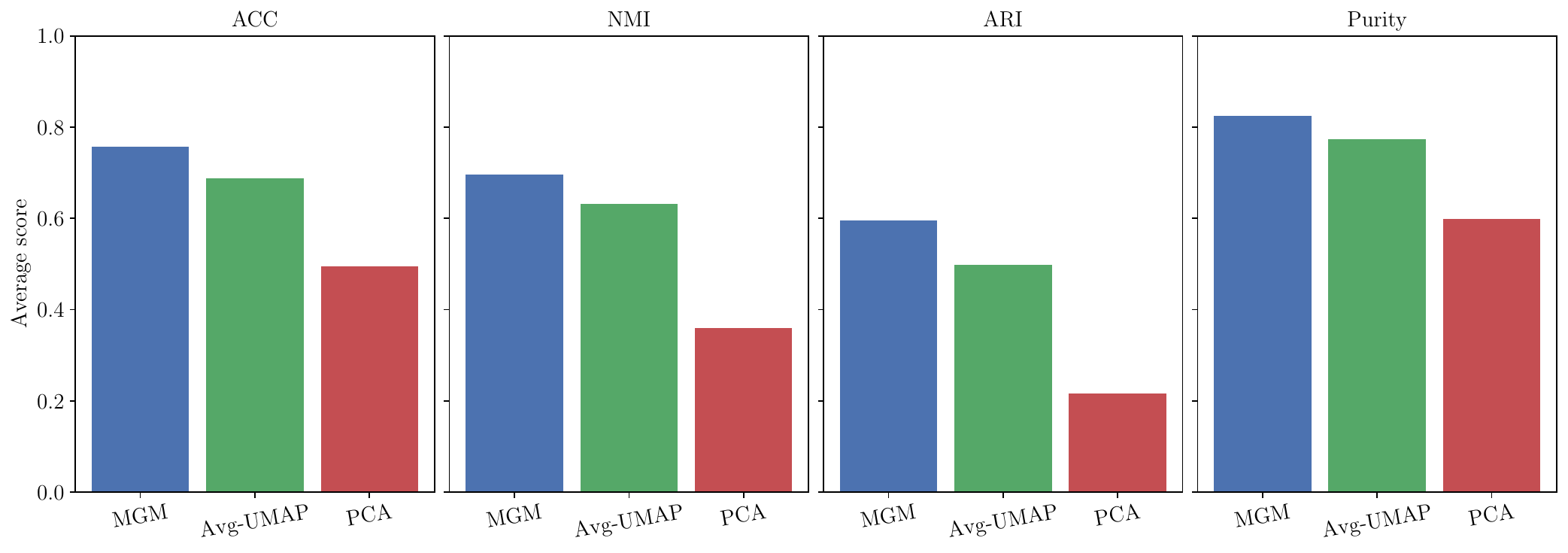}
  \caption{
Average clustering performance across all datasets under the noisy condition (Setup~I), 
comparing MGM, Avg-UMAP, and PCA. 
Panels correspond to four evaluation metrics: ACC, NMI, ARI, and Purity.
}
  \label{fig:avg_noisy}
\end{figure}

To better illustrate the overall trends across all datasets, 
Figure~\ref{fig:avg_noisy} presents the average values of the four metrics, 
providing an intuitive comparison of the three methods under the noisy setup.

\noindent\textbf{Analysis.}  
Figure~\ref{fig:avg_noisy} summarizes the average performance of the compared methods across all nine benchmark datasets.  
The proposed multiscale Grassmann manifolds (MGM) framework consistently ranks first or second across all four evaluation metrics, 
demonstrating greater stability than both  Avg-UMAP and PCA under noisy conditions.  
MGM achieves the best performance on most datasets and maintains reliable results across all metrics, 
whereas PCA often shows substantial degradation in the presence of noise and  Avg-UMAP exhibits higher variability across datasets.  
For medium-sized datasets such as GSE75748cell and GSE67835, 
the performance gap between MGM and the baselines is particularly pronounced, 
suggesting that the manifold-based representation more effectively preserves global and local structure when noise perturbs neighborhood geometry.  
Although  Avg-UMAP slightly outperforms MGM in a few isolated cases 
(e.g., GSE94820 for ACC and GSE84133human1 for Purity), 
MGM provides the most balanced and consistent performance overall.  
Notably, on the smallest dataset, GSE57249, MGM attains the highest scores across all metrics, 
demonstrating strong robustness even under extreme sample sparsity.  
These findings confirm that aggregating multiscale information within a Grassmann manifold subspace 
yields a more noise-resilient and geometrically stable representation than single-scale embeddings.

\subsubsection{Results under the refined condition (Setup II)}

\begin{figure}[!htbp]
  \centering
  \includegraphics[width=1\linewidth]{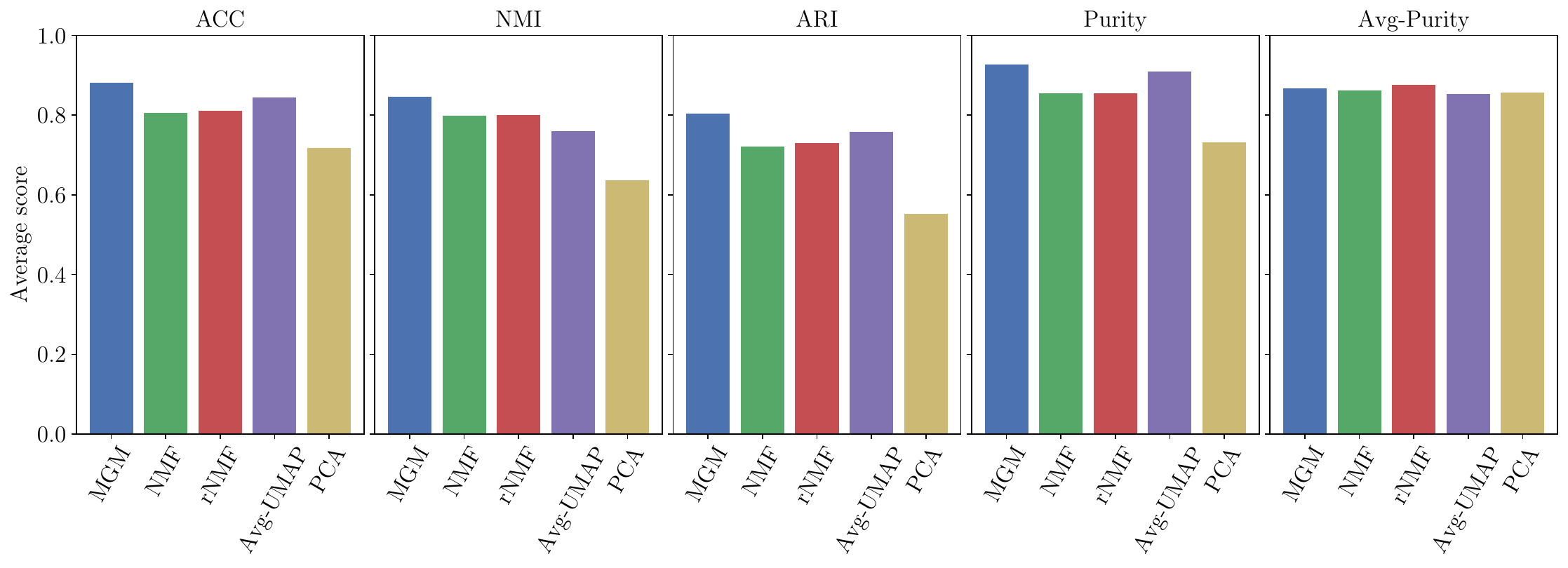}
  \caption{
Average clustering performance under the refined condition (Setup~II), 
comparing MGM, NMF, rNMF, Avg-UMAP, and PCA across five evaluation metrics 
(ACC, NMI, ARI, Purity, and Avg-Purity). 
Each panel shows the mean score of one metric across all datasets.
}
  \label{fig:setup2_avgs}
\end{figure}

\begin{figure*}[t]
  \centering
  \includegraphics[width=1\textwidth]{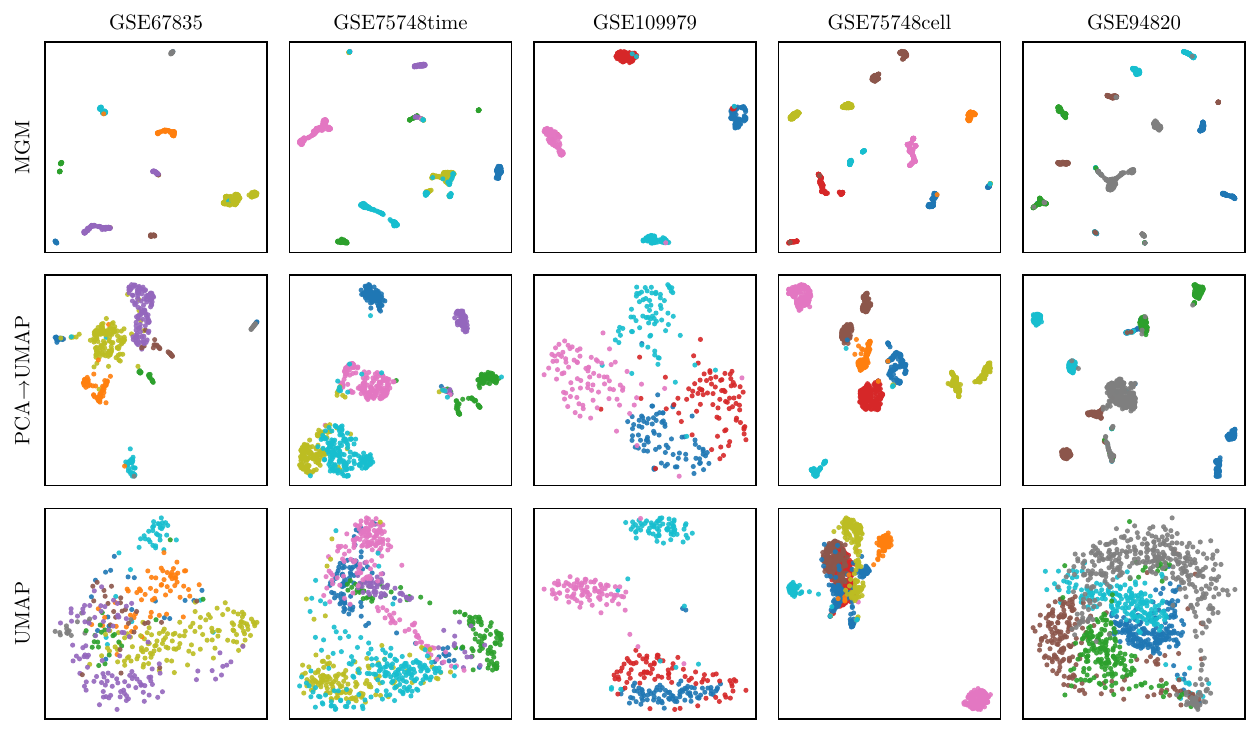}
  \caption{Visualization under UMAP (MGM vs.\ PCA$\rightarrow$UMAP vs.\ UMAP). 
Columns correspond to datasets (GSE67835, GSE75748time, GSE109979, GSE75748cell, GSE94820), and rows correspond to methods. 
Top row: MGM, where  chordal distances are computed from multiscale subspace representations and then embedded using UMAP with a precomputed metric. 
Middle row: PCA$\rightarrow$UMAP, obtained by applying UMAP to PCA-reduced features using the Euclidean metric. 
Bottom row: UMAP, obtained directly from the preprocessed data without PCA or Grassmann manifold representation.}
  \label{fig:panel_grumap}
\end{figure*}

Under the refined setting, five methods were compared: PCA, NMF, rNMF,  Avg-UMAP, and the proposed multiscale Grassmann manifolds (MGM) framework.  
With more rigorous preprocessing that reduces background noise and enhances data structure,  
MGM achieved the highest or comparable scores across most datasets and evaluation metrics.  
NMF and rNMF performed competitively on a few large-scale datasets (e.g., GSE84133Human1 and GSE84133Human2),  
reflecting its strength in well-conditioned low-rank representations.  
Detailed results for each metric are presented in Tables~\ref{tab:acc_refined}--\ref{tab:avgpurity_refined},  
followed by averaged performance across all datasets in Figure~\ref{fig:setup2_avgs}  
and qualitative visualization results in Figure~\ref{fig:panel_grumap}.

\begin{table}[H]
\centering
\caption{ACC under the refined condition (Setup II).  Best in \textbf{bold}.}
\label{tab:acc_refined}
\begin{tabular}{lccccc}
\hline
Dataset & MGM & NMF & rNMF & Avg-UMAP & PCA \\
\hline
GSE75748time        & \textbf{0.8776} & 0.6875\cite{hozumi2024analyzing} & 0.6873 \cite{hozumi2024analyzing} & 0.8618 & 0.8327 \\
GSE94820            & \textbf{0.9093} & 0.7189 \cite{hozumi2024analyzing} & 0.7091 \cite{hozumi2024analyzing} & 0.8598 & 0.5591 \\
GSE67835            & \textbf{0.8843} & 0.8364\cite{hozumi2024analyzing} & 0.8357 \cite{hozumi2024analyzing} & 0.7977 & 0.7390 \\
GSE75748cell        & \textbf{0.9851} & 0.8006  & 0.8045   & 0.9693 & 0.8686 \\
GSE109979           & \textbf{0.9780} & 0.7073  & 0.7073   & 0.9711 & 0.9311 \\
GSE57249            & \textbf{0.9796} & \textbf{0.9796} \cite{hozumi2024analyzing} & \textbf{0.9796} \cite{hozumi2024analyzing} & 0.9642 & 0.5510 \\
GSE84133human1      & 0.7490 & 0.7370 \cite{hozumi2024analyzing} & \textbf{0.7988} \cite{hozumi2024analyzing} & 0.6787 & 0.6451 \\
GSE84133human2      & 0.6860 & 0.8994 \cite{hozumi2024analyzing} & \textbf{0.8998} \cite{hozumi2024analyzing} & 0.6609 & 0.7073 \\
GSE84133human4      & \textbf{0.8910} & 0.8847 \cite{hozumi2024analyzing} & 0.8847 \cite{hozumi2024analyzing} & 0.8360 & 0.6431 \\
\hline
Average ACC& \textbf{0.8822} &	0.8057 &	0.8117	& 0.8444	 & 0.7170\\
\hline
\end{tabular}
\end{table}

\begin{table}[H]
\centering
\caption{NMI under the refined condition (Setup II). Best in \textbf{bold}.}
\label{tab:nmi_refined}
\begin{tabular}{lccccc}
\hline
Dataset & MGM & NMF & rNMF & Avg-UMAP & PCA \\
\hline
GSE75748time       & \textbf{0.8062} & 0.7244 \cite{hozumi2024analyzing} & 0.7227 \cite{hozumi2024analyzing} & 0.7958 & 0.8657 \\
GSE94820           & \textbf{0.7862} & 0.6693 \cite{hozumi2024analyzing} & 0.6624 \cite{hozumi2024analyzing} & 0.7441 & 0.4859 \\
GSE67835           & \textbf{0.8574} & 0.8017 \cite{hozumi2024analyzing} & 0.7975 \cite{hozumi2024analyzing} & 0.8063 & 0.6814 \\
GSE75748cell       & \textbf{0.9592} & 0.8854   & 0.8897   & 0.9468 & 0.9229 \\
GSE109979\_329cell & \textbf{0.9219} & 0.6321   & 0.6321   & 0.9003 & 0.8831 \\
GSE57249           & \textbf{0.9293} & \textbf{0.9293 }\cite{hozumi2024analyzing} &\textbf{ 0.9293} \cite{hozumi2024analyzing} & 0.4449 & 0.2270 \\
GSE84133human1     & 0.7881 & 0.7949 \cite{hozumi2024analyzing} & \textbf{0.8226} \cite{hozumi2024analyzing} & 0.6773 & 0.5336 \\
GSE84133human2     & 0.7519 & 0.8829 \cite{hozumi2024analyzing} & \textbf{0.8835} \cite{hozumi2024analyzing} & 0.7361 & 0.6347 \\
GSE84133human4     & 0.8195 & \textbf{0.8694} \cite{hozumi2024analyzing} & \textbf{0.8694} \cite{hozumi2024analyzing} & 0.7949 & 0.4914 \\
\hline
Average NMI& \textbf{0.8466}&	0.7988&	0.8010&	0.7607&	0.6362\\
\hline
\end{tabular}
\end{table}

\begin{table}[h]
\centering
\caption{ARI under the refined condition (Setup II). Best in \textbf{bold}.}
\label{tab:ari_refined}
\begin{tabular}{lccccc}
\hline
Dataset & MGM & NMF & rNMF & Avg-UMAP & PCA \\
\hline
GSE75748time        & 0.7350 & 0.5996\cite{hozumi2024analyzing} & 0.5969\cite{hozumi2024analyzing} & 0.7203 & \textbf{0.7470} \\
GSE94820            & \textbf{0.7942} & 0.5556\cite{hozumi2024analyzing} & 0.5440\cite{hozumi2024analyzing} & 0.7317 & 0.3365 \\
GSE67835            & \textbf{0.8214} & 0.7314\cite{hozumi2024analyzing} & 0.7295\cite{hozumi2024analyzing} & 0.7190 & 0.6253 \\
GSE75748cell        & \textbf{0.9675} & 0.7479 & 0.7512 & 0.9403 & 0.8171 \\
GSE109979  & \textbf{0.9428} & 0.5727 & 0.5727 & 0.9253 & 0.8834 \\
GSE57249            & \textbf{0.9483} & \textbf{0.9483}\cite{hozumi2024analyzing} & \textbf{0.9483}\cite{hozumi2024analyzing} & 0.9145 & 0.1438 \\
GSE84133human1      & 0.5987 & 0.6120\cite{hozumi2024analyzing} & 0.7080\cite{hozumi2024analyzing} & 0.5386 & 0.4289 \\
GSE84133human2      & 0.5961 & \textbf{0.8929}\cite{hozumi2024analyzing} & 0.8930\cite{hozumi2024analyzing} & 0.5684 & 0.5819 \\
GSE84133human4      & 0.8245 & \textbf{0.8311}\cite{hozumi2024analyzing} & \textbf{0.8311}\cite{hozumi2024analyzing} & 0.7638 & 0.4031 \\
\hline
Average ARI& \textbf{0.8032}	&0.7213	&0.7305	& 0.7580 &	0.5519\\
\hline
\end{tabular}
\end{table}

\begin{table}[h]
\centering
\caption{Purity under the refined condition (Setup II). Best in \textbf{bold}.}
\label{tab:purity_refined}
\begin{tabular}{lccccc}
\hline
Dataset & MGM & NMF & rNMF & Avg-UMAP & PCA \\
\hline
GSE75748time     & \textbf{0.8776} & 0.7455\cite{hozumi2024analyzing} & 0.7467\cite{hozumi2024analyzing} & 0.8634 & 0.8427 \\
GSE94820         & \textbf{0.9093} & 0.7531\cite{hozumi2024analyzing} & 0.7429\cite{hozumi2024analyzing} & 0.8643 & 0.5607 \\
GSE67835         & \textbf{0.9052} & 0.8719\cite{hozumi2024analyzing} & 0.8726\cite{hozumi2024analyzing} & 0.8748 & 0.7490 \\
GSE75748cell     & \textbf{0.9851} & 0.8301 & 0.8310 & 0.9696 & 0.9016 \\
GSE109979       & \textbf{0.9780} & 0.7073 & 0.7073 & 0.9711 & 0.9311 \\
GSE57249         & \textbf{0.9796} & \textbf{0.9796}\cite{hozumi2024analyzing} & \textbf{0.9796}\cite{hozumi2024analyzing} & 0.9642 & 0.5755 \\
GSE84133human1   & 0.8941 & 0.9099\cite{hozumi2024analyzing} & \textbf{0.9189}\cite{hozumi2024analyzing} & 0.8813 & 0.6528 \\
GSE84133human2   & 0.9153 & 0.9600\cite{hozumi2024analyzing} & \textbf{0.9602}\cite{hozumi2024analyzing} & 0.9040 & 0.7291 \\
GSE84133human4   & 0.9070 & 0.9412\cite{hozumi2024analyzing} & \textbf{0.9412}\cite{hozumi2024analyzing} & 0.8870 & 0.6449 \\
\hline
Average Purity& \textbf{0.9279}&	0.8554 &	0.8556&	0.9089 & 0.7319\\
\hline
\end{tabular}
\end{table}

\begin{table}[!htbp]
\centering
\caption{Avg-Purity under the refined condition (Setup II). 
Best in \textbf{bold}. }
\label{tab:avgpurity_refined}
\begin{tabular}{lccccc}
\hline
Dataset & MGM & NMF & rNMF & Avg-UMAP & PCA \\
\hline
GSE75748time          & 0.8546 & 0.8140  & 0.8138  & 0.8440 & \textbf{0.9116} \\
GSE94820              & 0.8383 & 0.8211  & 0.8047 & 0.7981 & \textbf{0.8445} \\
GSE67835              & 0.8452 & 0.7859  & 0.7409  & 0.7989 & \textbf{0.8819} \\
GSE75748cell          & \textbf{0.9637} & 0.9260 & 0.9310 & 0.9519 & 0.9382 \\
GSE109979    & \textbf{0.9557} & 0.8497 & 0.8497 & 0.9420 & 0.9459 \\
GSE57249              & \textbf{0.9342} & 0.9342  & 0.9342  & 0.9154 & 0.6624 \\
GSE84133human1        & \textbf{0.8685} & 0.8576  & 0.8584 & 0.8412 & 0.8583 \\
GSE84133human2        & 0.7717 & 0.8897  & \textbf{0.8990}  & 0.7508 & 0.8052 \\
GSE84133human4        & 0.8594 & \textbf{0.9238} & \textbf{0.9238}  & 0.8403 & 0.8656 \\
\hline
Average Avg-Purity&  0.8669 &	0.8617&	\textbf{0.8768}&	0.8536& 0.8571\\
\hline
\end{tabular}
\end{table}
\noindent
To consolidate these per-metric outcomes, Figure~\ref{fig:setup2_avgs} reports the averages across the nine datasets for each metric, offering an at-a-glance comparison of the five methods.

\begin{remark}
Results for NMF and rNMF under the metrics of ACC, NMI, ARI, and Purity 
are adopted from~\cite{hozumi2024analyzing}, 
except for the datasets \textit{GSE75748cell} and \textit{GSE109979}, 
which were reproduced using the authors’ publicly available code, 
as these datasets were not reported in the original paper. 
For Avg-Purity, all results were independently computed in this study 
using the same publicly available implementation for consistency. 
For fair comparison, PCA results were obtained by performing clustering 
on embeddings reduced to the same dimensionality as that used in the proposed MGM framework.
\end{remark}

\noindent
\textbf{Analysis.}  
Across the five evaluation metrics: ACC, NMI, ARI, purity, and Avg-Purity, the proposed multiscale Grassmann manifolds (MGM) framework demonstrates strong and consistent performance under the refined condition (Setup~II). 
As summarized in Tables~\ref{tab:acc_refined}–\ref{tab:avgpurity_refined}, MGM achieves the highest overall mean values on four of the five metrics (ACC, NMI, ARI, and Purity). 
For Avg-Purity, MGM ranks second overall, only slightly below rNMF, with a marginal difference. 
This consistency across all metrics indicates that integrating multiscale geometric information within the Grassmann manifold yields stable and competitive clustering representations even after refined preprocessing. A closer examination of the dataset-wise results shows that MGM achieves the best or comparable performance on the majority of datasets, particularly those of small and medium scale, including GSE75748time, GSE94820, GSE67835, GSE75748cell, GSE109979, and GSE57249. 
In these datasets, MGM consistently attains top scores across multiple metrics, highlighting its advantage in preserving multiscale structure and local manifold geometry. 
This suggests that the subspace-based representation on the Grassmann manifold is especially effective when the number of cells is limited or residual noise remains after preprocessing. 
For the larger datasets (GSE84133human1, GSE84133human2, and GSE84133human4), NMF slightly surpasses MGM on certain metrics, likely because factorization-based methods capture broader global relationships in complex, high-dimensional data. 
Nevertheless, MGM still outperforms PCA and Avg-UMAP across all metrics and remains highly competitive with NMF and rNMF overall.

Finally, to complement the quantitative evaluation, 
Figure~\ref{fig:panel_grumap} presents a direct visual comparison of the embeddings generated by the proposed multiscale Grassmann manifolds (MGM) framework, 
PCA$\rightarrow$UMAP (i.e., applying UMAP on features that have been first reduced in dimension using PCA), 
and standard UMAP computed directly on the preprocessed data.

\noindent\textbf{Observation.}  
In Figure~\ref{fig:panel_grumap}, across the five representative datasets, the MGM embeddings consistently produce clearer inter-cluster separation and more compact group structures compared with PCA$\rightarrow$UMAP and UMAP.  
In particular, the MGM results for GSE75748cell and GSE109979 show well-defined and isolated clusters, whereas PCA$\rightarrow$UMAP retains partial overlap between groups and UMAP exhibits widespread dispersion with blurred boundaries.  
For GSE67835 and GSE75748time, MGM preserves global organization while maintaining distinct local partitions, indicating that the Grassmann manifold representation enhances multiscale consistency.  
On GSE94820, MGM better preserves between-group separation, whereas PCA and UMAP fail to disentangle nearby populations.  
Overall, these visual results confirm the quantitative findings in Figure~\ref{fig:setup2_avgs}, showing that the MGM framework provides the most balanced representation, effectively enhancing separability while maintaining stable global structure across diverse single-cell datasets.

\begin{remark}
All visualizations in Figure~\ref{fig:panel_grumap} are generated under the parameter settings of Setup~II. 
Both MGM and PCA$\rightarrow$UMAP are constructed using features reduced to 50 dimensions by PCA, 
ensuring that the comparison focuses on the effect of the representation framework rather than differences in input dimensionality. 
The neighborhood size in UMAP is set to its default value of 15.
\end{remark}

\section{Discussion}
\label{sec:discussion}

The experimental results in Section~4 demonstrate that the proposed multiscale Grassmann manifolds (MGM) framework provides stable
and competitive clustering performance across a broad range of single-cell
datasets. It performs particularly well on small and moderately sized datasets,
where it is competitive with NMF, while maintaining comparable performance on
larger datasets. To further clarify the behavior of MGM under different scale
conditions, two representative examples are examined in the Supporting
Information (Section~2.1 and Section~2.2).

Section~2.1 in the Supporting Information presents the GSE57249 dataset,
where all scales yield consistently high-quality single-scale
UMAP embeddings. In this uniform condition, MGM preserves the discriminative
structure without degradation and maintains results comparable to the best
single-scale configuration, slightly improving over the average UMAP score.
This observation suggests that aggregating subspaces on the Grassmann manifold does
not blur or weaken well-formed local relationships when the underlying
scales are uniformly strong.

Section~2.2 in the Supporting Information presents the GSE84133Human2 dataset,
where different scales produce variable UMAP embeddings, and some
large-scale settings are affected by noise and over-smoothing. In this noisy
condition, MGM combines information from both effective and weak scales to
form a more stable subspace representation. It shows higher accuracy and
information-based metrics than the average UMAP result and sometimes performs
better than the best single-scale case. This suggests that Grassmann manifold
integration helps maintain robustness when individual scales do not perform
well.

Together, the two cases highlight the essential advantage of the proposed
framework. MGM can retain discriminative information under uniformly
good scales while compensating for instability under noisy or heterogeneous
conditions. This dual robustness underscores the potential of multiscale
Grassmann manifold learning. Although the present implementation relies on a simple
power-based sampling of scale sizes, the observed improvements suggest
that more adaptive or data-driven scale selection strategies could further
strengthen its capability in handling diverse single-cell datasets.

In this study, we demonstrated the effectiveness of the proposed multiscale Grassmann manifolds (MGM) framework using UMAP as a representative dimensionality reduction method. 
Nevertheless, the framework itself is general and can be readily combined with a wide range of existing algorithms, 
including t-SNE~\cite{maaten2008visualizing}, NMF~\cite{lee1999learning, feng2024multiscale}, 
MCIST~\cite{cottrell2025multiscale}, and CCP-assisted methods~\cite{hozumi2024analyzing}. 
By providing a unified geometric formulation for integrating multiscale information, 
the MGM framework establishes a flexible foundation that can be extended to other manifold learning or matrix factorization models. 
We anticipate that future studies will explore these directions, 
further advancing the application of Grassmann manifold representations in single-cell data analysis and beyond.

\section{Conclusion}

The proposed framework is designed to be general rather than specific to a single algorithm. 
While UMAP was used here to illustrate its feasibility, the multiscale Grassmann manifolds (MGM) formulation can be seamlessly integrated with other dimensionality reduction and representation learning methods, 
such as t-SNE~\cite{maaten2008visualizing}, NMF~\cite{lee1999learning, hozumi2024analyzing}, 
MCIST~\cite{cottrell2025multiscale}, and CCP-assisted approaches~\cite{hozumi2024analyzingccp}. 
By representing multiscale embeddings as points on the Grassmann manifold, the framework captures structural consistency across scales 
and provides a unified non-Euclidean representation for complex, high-dimensional data. 
Experiments on nine benchmark single-cell RNA-seq datasets demonstrated that MGM achieves competitive  clustering performance 
compared with conventional Euclidean and manifold-based approaches, particularly on small to medium-sized datasets where scale variability and noise have stronger effects. 

Future research may extend this framework toward adaptive or data-driven scale selection 
and more efficient subspace construction to improve robustness and scalability on large or heterogeneous datasets. 
Beyond single-cell analysis, the same formulation can be applied to other high-dimensional domains, 
offering a general pathway for integrating multiscale information through geometric modeling. 
Overall, the MGM framework establishes a principled and extensible foundation for multiscale geometric learning, 
supporting broader applications of Grassmann manifold representations in biological and computational data science.
 
\section*{Data and Code Availability}

All source code is available at \url{https://github.com/XiangXiangJY/MGM},  and the single-cell dataset preprocessing scripts are adapted from  
\url{https://github.com/hozumiyu/TopologicalNMF-scRNAseq}.

 \section*{Supporting Information}

Additional materials, including 
complete parameter configurations for Setup~I and Setup~II, supplementary
experimental analyzes, clustering settings, and detailed definitions of the evaluation metrics, are provided in the
Supporting Information at \url{https://github.com/XiangXiangJY/MGM}.

	    \section*{Acknowledgments}
This work was supported in part by NIH grant R35GM148196, National Science Foundation grant DMS2052983,  Michigan State University Research Foundation, and  Bristol-Myers Squibb 65109.

%\bibliographystyle{unsrt}
% \bibliographystyle{unsrt}
%\myexternaldocument{supplementary}
\bibliographystyle{plain}
\bibliography{refs}
   
\end{document}